# Exploiting System Hierarchy to Compute Repair Plans in Probabilistic Model-Based Diagnosis

**Sampath Srinivas**
Computer Science Department
Stanford University
Stanford, CA 94305
srinivas@cs.stanford.edu

**Eric Horvitz**
Decision Theory Group
Microsoft Research
Redmond, WA 98052
horvitz@microsoft.com

## Abstract

The goal of model-based diagnosis is to isolate causes of anomalous system behavior and recommend cost-effective repair actions. In general, precomputing optimal repair policies is intractable. To date, investigators addressing this problem have explored approximations that either impose restrictions on the system model, such as a single fault assumption, or that compute an immediate best action with limited lookahead. In this paper, we develop a formulation of repair in model-based diagnosis and a repair algorithm that computes optimal sequences of actions. This optimal approach is costly but can be applied to precompute an optimal repair strategy for compact systems. We show how we can exploit a hierarchical system specification to make this approach tractable for larger systems. When introducing hierarchy, we also consider the tradeoff between simply replacing a component and decomposing it to repair its subcomponents. The hierarchical repair algorithm is suitable for off-line precomputation of an optimal repair strategy. A modification of the algorithm takes advantage of an iterative-deepening scheme to trade off inference time and the quality of the computed strategy.

## 1 INTRODUCTION

The goal of probabilistic model-based diagnosis is computation of a minimum expected cost sequence of observation and repair actions to restore a malfunctioning system to working order. In general, the computation of such an optimal repair strategy is intractable. We must consider each possible strategy in a combinatorial space of repair strategies to compute the optimal strategy.

The intractability of general diagnosis can be addressed in several ways. Several researchers have explored methods that rely on the imposition of a greedy, myopic assumption for generation of repair strategies (for example, [Friedrich & Nejdl, 1992; Heckerman et al, 1995]). These methods compute a good immediate repair action or policy. A best next action is carried out, revealing additional information about the world which is used to compute a revised policy. Investigators have also imposed restrictions on the system model. A common restriction is the assumption that only a single fault is present [Kalagnanam & Henrion, 1988; Heckerman et al, 1995].

In this paper, we take a different approach. Rather than relaxing the goal of optimality via assumptions of greediness or imposition of modeling restrictions, we gain tractability by exploiting the hierarchical structure of complex systems. The rest of the paper is structured as follows: First, we formalize the general repair problem in non-hierarchical systems and develop an algorithm to generate optimal repair policies. The initial algorithm we develop is only suitable for systems with small numbers of components because the problem it solves is inherently exponential in the number of components. We show how we can extend the applicability of the approach by adapting the optimal repair algorithm to handling systems cast as a hierarchy of components. This results in an algorithm that can be used off-line to tractably compute optimal repair policies for large hierarchical systems. We show how the algorithm exploits the hierarchy to gain tractability. We then introduce flexibility into the algorithm so that it can be used in real time. We do so by incorporating an iterative-deepening scheme that trades off the timeliness of a response for the quality of the computed strategy. We conclude with a discussion examining related work.

## 2 PROBLEM DEFINITION

Consider a system with $n$ components, any of which may fail. Each component $C_i$ has a set of discrete-valued inputs $I_i^1, I_i^2, \ldots, I_i^k$ and one discrete-valued output $O_i$. We refer to the vector of variables $\langle I_i^1, I_i^2, \ldots, I_i^k \rangle$ as $\mathbf{I}_i$.

Let us assume that the outputs and inputs of the components are connected in accordance with the causal flow of a system—we assume that there are no feedback loops in the system. We also assume that the system has a single designated output variable. Hence, when viewed as a black box, the system has a set of input variables which we will call the *system input* variables and one output variable which we will call the *system output* variable.

Each component $C_i$ can assume any of a set of possible operating states. One of these states $ok_i$ is a distinguished state which corresponds to the component's *normal mode* of operation. The other states of $C_i$ correspond to states where the component is not operating normally. We will assume that components fail independently. The state of $C_i$ is modeled by a probabilistic state variable $M_i$. The user specifies a prior probability distribution over the states of $M_i$. This distribution quantifies the reliability of $C_i$ and is specified based on empirical data and expert knowledge.

The model of operation of the component specifies the value of the output of the component given the state of the component and the values of the inputs. When the component $C_i$ is in the state $ok_i$ we will assume that a deterministic model of operation is available – that is, given an input state, exactly one output state is possible. However, for the other states, we will leave open the possibility that the behavior is non-deterministic; for each of these states, the user can specify a probability distribution over the output for every possible input state. Specifying the model of operation of the component amounts to specifying the distribution $P(O_i|M_i, I_i)$ (with the restriction that $P(O_i = o_i|M_i = ok_i, I_i = i_i)$ always takes the value 0 or 1 for any $o_i$ and $i_i$). Finally, the user specifies a cost of replacement $c_i$ for $C_i$. $c_i$ is the materials and labor cost of simply discarding the existing part and replacing it with a new one.

Say we are observing the artifact modeled by our system model and we observe some anomaly. That is, the system is given an input vector (which we can observe) and we observe that the system output is inconsistent with the correct operation of the system.

We wish to take actions to repair the system. We will define repairing the system as correcting the *perceived anomalous observation*. The actions available to us are replacement of components. A *repair strategy* is a sequence in which the components are replaced. After each successive replacement we check the output of the system. We assume the input is fixed to the input vector associated with the observed anomaly. If the output is still anomalous, we continue to replace the next component recommended by the strategy. If the output is no longer anomalous we stop. Executing each possible repair strategy (*i.e.*, each repair sequence) has an expected cost. The optimal repair strategy is the one with the least expected cost for a particular system input.

If we compute the optimal strategy for each possible system input value, we determine the *optimal repair plan* for the system. The optimal repair plan is a set of situation–action rules. If a system has anomalous behavior, the optimal repair plan gives us a repair strategy to use as a function of the input. We will now develop an algorithm to compute optimal repair plans.

## 2.1 COMPUTING THE OPTIMAL REPAIR PLAN

Consider a system which has a vector of input variables **I**. Let the system output variable be $X$. Let us assume that the system has been given an input **i**. Further, we assume that the *correct* system output for this input is $x(\mathbf{i})$ and we have observed that the output $X$ has some value other than the correct output[1].

Consider a repair strategy $R = \langle C_1, C_2, \ldots, C_n \rangle$. $R$ is a sequence describing the order in which components will be replaced. We now develop an expression for the expected cost of $R$. We will refer to the action of replacing $C_i$ as $fix_i$. In addition, we will refer to the system output $X$ after replacement of the $i$-th component in the sequence as $X_i$. Note in particular that the variable $X_0$ denotes the value of $X$ before replacement of any component. Let $S_j$ be the sequence of observations and actions up to and including the replacement of $C_j$. That is: $S_j = [\mathbf{I} = \mathbf{i}, X_0 = \neg x(\mathbf{i}), fix_1, X_1 = \neg x(\mathbf{i}), fix_2, X_2 = \neg x(\mathbf{i}), \ldots, fix_{j-1}, X_{j-1} = \neg x(\mathbf{i}), fix_j]$. The expected cost of $R$ is given by:

$$EC(R|\mathbf{I} = \mathbf{i}, X = \neg x(\mathbf{i})) = \qquad (1)$$
$$(c_1 +$$
$$P(X_1 = \neg x(\mathbf{i})|S_1)(c_2 +$$
$$P(X_2 = \neg x(\mathbf{i})|S_2)(c_3 +$$
$$\ldots$$
$$P(X_j = \neg x(\mathbf{i})|S_j)(c_{j+1}$$
$$\ldots$$
$$P(X_{n-1} = \neg x(\mathbf{i})|S_{n-1})c_n \ldots) \ldots)))$$

To determine the expected cost as described in Equation 1, we need to compute the conditional probabilities $P(X_j = \neg x(\mathbf{i})|S_j)$, $1 \leq j < n$.

We demonstrate how these probabilities can be computed using an example. Consider the electronic circuit in Fig 1(a). This system model for this circuit can be translated into a Bayesian network as displayed in Fig 1(b) (see [Srinivas, 1994] for details about this transformation). We will elucidate the computation of the conditional distribution using this Bayesian network. Consider the repair sequence $R = \langle AND, XOR, OR \rangle$. The network of Fig 2 represents the situation after the $AND$ gate has been replaced. The modes of the $XOR$ and $OR$ gates are unaffected by this replacement and have the same value both before and after the repair action. The arcs between the copies of the static network in Fig 2 model this persistence[2].

---

[1]Note that we can determine the correct system output by simply simulating the system forward from the input **i** while assuming that each of the components $C_i$ are in the $ok_i$ state.

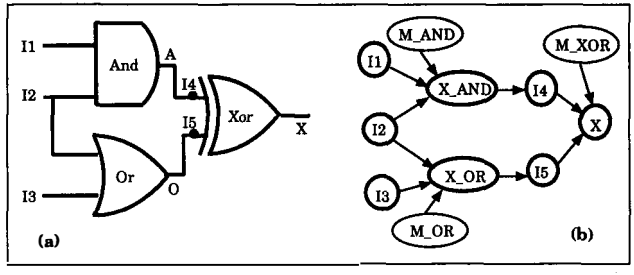

Figure 1: An electronic-circuit example. (a) A system model showing components and interconnectivity, and (b) the corresponding Bayesian network for the system model.

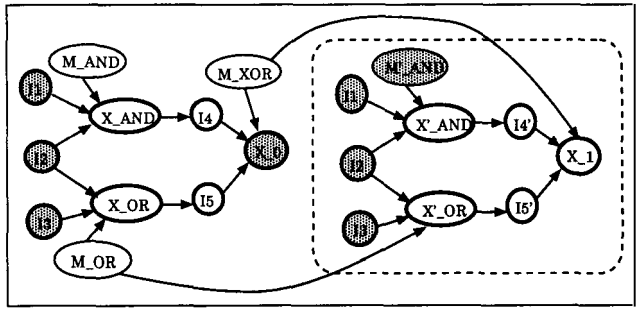

Figure 2: Representation of the situation after replacing the $AND$ gate. Arcs between copies of the static network represent persistence of state.

The probability $P(X_1 = \neg x(\mathbf{i})|S_1)$ can be computed by declaring the evidence $S_1$ in the network, propagating it and then looking up the posterior belief of the event $X_1 = \neg x(\mathbf{i})$ in node $X_1$. The evidence $S_1$ consists of: (a) the known state of the input both before and after the repair action, (b) the output $X_0$ (which has the value $\neg x(\mathbf{i})$), and (c) the state of the $AND$ gate after repair ($M'_{AND} = \mathbf{ok}$). The corresponding nodes are shown shaded gray in the figure.

We will now develop an iterative scheme for making this computation without explicitly constructing the dynamic Bayesian network of Fig 3. We begin by describing one step of this iterative scheme.

We note that there are *active paths* (see [Pearl, 1988])

---

[2] See [Balke & Pearl, 1994; Heckerman & Shachter, 1994; Darwiche & Goldszmidt, 1994] for details on modeling persistence.

to node $X_1$ from node $X_0$ through the nodes $M_{XOR}$ and $M_{OR}$. Hence the computation of the posterior of $X_1$ will necessarily have to consider cases for every possible joint state[3] of the variables $M_{XOR}$ and $M_{OR}$. The computation can be written as follows:

$$P(X_1 = \neg x(\mathbf{i})|S_1) =$$
$$\Sigma_{m_{XOR}, m_{OR}} P(X_1 = \neg x(\mathbf{i})|\mathbf{I} = \mathbf{i}, M'_{AND} = \mathbf{ok},$$
$$M_{XOR} = m_{XOR}, M_{OR} = m_{OR}) \times$$
$$P(M_{XOR} = m_{XOR}, M_{OR} = m_{OR}|$$
$$\mathbf{I} = \mathbf{i}, X_0 = \neg x(\mathbf{i})) \quad (2)$$

In the above equation, $m_{XOR}$ and $m_{OR}$ represent generic states of $M_{XOR}$ and $M_{OR}$ respectively. Hence the summation in the equation iterates over all possible joint states of $M_{XOR}$ and $M_{OR}$. The equation also accounts for the fact that knowing the state of $M_{XOR}$ and $M_{OR}$ makes $X_1$ conditionally independent of $X_0$.

Assume that we have access to the probability distribution $P(M_{XOR}, M_{OR}|\mathbf{I} = \mathbf{i}, X_0 = \neg x(\mathbf{i}))$ (*i.e.*, the second term of Equation 2). As we shall see later, this distribution is the output of the previous step of the iterative scheme. We note that the probabilities needed for the first term of the equation, $P(X_1 = \neg x(\mathbf{i})|\mathbf{I} = \mathbf{i}, M'_{AND} = \mathbf{ok}, M_{XOR} = m_{XOR}, M_{OR} = m_{OR})$, can be computed directly from the static Bayesian network of Fig 1(b). This is so because knowing the states of $M_{XOR}$ and $M_{OR}$ makes the post-repair network fragment (shown within the dotted lines in Fig 2) independent of the rest of the network. The post-repair network fragment is identical to the static Bayesian network. Thus, the required probability can be computed by (a) declaring the evidence $M'_{AND} = \mathbf{ok}$, $M_{XOR} = m_{XOR}$, $M_{OR} = m_{OR}$, $\mathbf{I} = \mathbf{i}$ in the static network, and (b) propagating the evidence and looking up the posterior of the event $X = \neg x(\mathbf{i})$ in node $X$.

Now, let us consider the replacement of the next component specified by the repair strategy, *i.e.*, the $XOR$ gate. The situation is shown in Fig 3. By analogy the previous situation, we can compute $P(X_2 = \neg x(\mathbf{i})|S_2)$ as:

$$P(X_2 = \neg x(\mathbf{i})|S_2) =$$
$$\Sigma_{m_{OR}} P(X_2 = \neg x(\mathbf{i})|\mathbf{I} = \mathbf{i}, M'_{XOR} = \mathbf{ok},$$
$$M'_{AND} = \mathbf{ok}, M_{OR} = m_{OR}) \times$$
$$P(M_{OR} = m_{OR}|\mathbf{I} = \mathbf{i}, X_0 = \neg x(\mathbf{i}),$$
$$M'_{AND} = \mathbf{ok}, X_1 = \neg x(\mathbf{i})) \quad (3)$$

Here, again, the probabilities for the first term in the equation can be directly computed from the system model. We will now see how the probabilities for the second term can be computed from the computations of the previous repair step (*i.e.*, Equation 2). Note that the product within the summation of Equation 2

---

[3] A joint state of a set of discrete random variables assigns a value to each of the variables in the set.

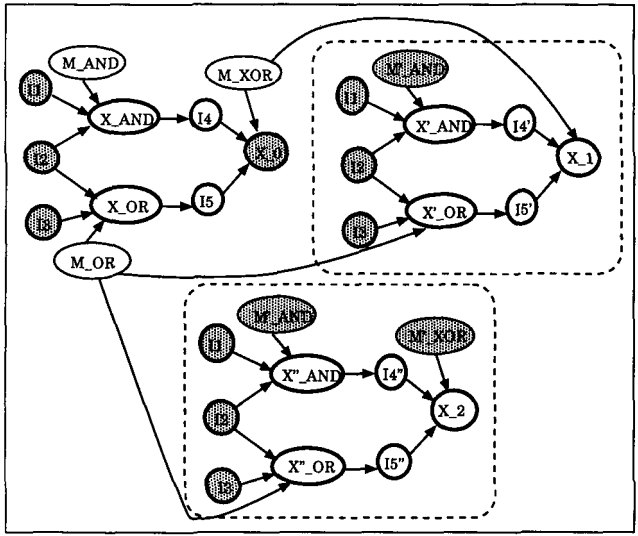

Figure 3: Situation after replacing the $XOR$ gate.

is equal to:

$$P(X_1 = \neg x(\mathbf{i}), M_{XOR} = m_{XOR}, M_{OR} = m_{OR}|$$
$$\mathbf{I} = \mathbf{i}, X_0 = \neg x(\mathbf{i}), M'_{AND} = \mathbf{ok})$$

If we sum the distribution above over all joint states of $M_{XOR}$ and $M_{OR}$ we obtain the marginal $P(X_1 = \neg x(\mathbf{i})|\mathbf{I} = \mathbf{i}, X_0 = \neg x(\mathbf{i}), M'_{AND} = \mathbf{ok})$. Dividing each element the distribution above by this marginal gives us:

$$P(M_{XOR} = m_{XOR}, M_{OR} = m_{OR}|\mathbf{I} = \mathbf{i},$$
$$X_0 = \neg x(\mathbf{i}), M'_{AND} = \mathbf{ok}, X_1 = \neg x(\mathbf{i})) \quad (4)$$

Now, we consider the action of replacing the $XOR$ gate. This does not affect our estimate of what state the $OR$ gate is in; it has no affect on the posterior probability distribution of $M_{OR}$ given the current state of information. This posterior probability is:

$$P(M_{OR} = m_{OR}|\mathbf{I} = \mathbf{i}, X_0 = \neg x(\mathbf{i}), M'_{AND} = \mathbf{ok},$$
$$X_1 = \neg x(\mathbf{i}))$$

This posterior distribution can be computed by summing the distribution of Equation 4 over all states of $M_{XOR}$. Note that this distribution is precisely what we need to solve Equation 3.

This example can be generalized to yield a simple iterative scheme for computing $P(X_j = \neg \mathbf{x}(\mathbf{i})|S_j)$ for $1 \leq j < n$, given a repair sequence $R$. The key idea is that all of the information coming from the first $j-1$ observation repair actions is summarized by the posterior probability distribution over the joint states of the components *which have not yet been fixed* (i.e., $\mathcal{C}_j$ through $\mathcal{C}_n$). This posterior probability is used iteratively to perform the following calculations:

- Compute the probability of an anomaly after the $j$-th fix action (i.e., $P(X_j = \neg \mathbf{x}(\mathbf{i})|S_j)$).

- Compute the new updated posterior, accounting for the $j$-th action. Note that this posterior is over the joint states of $\mathcal{C}_{j+1}$ through $\mathcal{C}_n$.

To begin the iteration, we need to have the posterior over all joint states of all the components, given that no observation and repair actions have been performed. Note that this posterior probability is just the prior probability over the joint states of the components. As components fail independently, this distribution is the product of the marginal distributions over the modes of each component.

In general, when computing $P(X_j = \neg x(\mathbf{i})|S_j)$ we note that there is an active path in the corresponding dynamic Bayesian network from the observed node $X_{j-1}$ to the target node $X_j$ through the mode variable of each unfixed component (i.e., there are active paths through $M_{j+1}, M_{j+2}, \ldots, M_n$). As a result, a cutset for the network necessarily includes each of these variables. This implies that any inference algorithm computing $P(X_j = \neg x(\mathbf{i})|S_j)$ will necessarily have to condition on each joint state of the modes of the unfixed components. Our iterative scheme explicitly does this conditioning by carrying forward the posterior over the cutset nodes. Thus, its complexity is of the same order as any other scheme for performing this computation. In this sense, our scheme is for computing the probabilities $P(X_j = \neg x(\mathbf{i})|S_j)$ is optimal.

### 2.1.1 Computing the best strategy

We now have the probabilities required to solve Equation 1, enabling us to compute the expected cost $EC(R|\mathbf{I} = \mathbf{i}, X = \neg x(\mathbf{i}))$, given a strategy $R$ and a system input $\mathbf{I} = \mathbf{i}$. Identifying the best strategy can be done by checking the expected cost of all strategies. Computing the best strategy can be done in $O(n!S_M)$ where $n$ is the number of components and $S_M$ is the joint space size of the mode variables of all the components. $S_M$ is exponential in $n$.

To compute the optimal repair plan, we must compute the best possible strategy for every possible input value to the system. Let the joint space size of the inputs to the system be $S_I$. The overall complexity of computing the optimal strategy for every possible input to the system is then $O(n! \times S_M \times S_I)$. Computing the optimal repair plan with the algorithm described above is impractical for large systems. We now modify our algorithm to exploit the hierarchical structure of systems so that the method can be scaled up to large systems.

## 3  HIERARCHICAL SYSTEMS

It is common in engineering practice to manage the complexity of systems by organizing them into hierarchical assemblies of components. Hierarchical configurations simplify the task of design, construction, and repair of complex systems by grouping system compo-nents into subsystems with well-defined functionalities and manageable interdependencies. We will now focus on the details of an extension to the general repair formulation and inference procedure that allows us to take advantage of hierarchical system specifications. The extension builds on a formulation of hierarchical systems in [Srinivas, 1994].

### 3.1 DEFINING THE HIERARCHICAL MODEL

When performing hierarchical modeling, an engineer has the option of modeling each component of the system either atomically or hierarchically. An atomic component model has no further substructure, and is defined exactly as in Section 2. A hierarchical component model specifies the behavior of components in terms of the behaviors of its subcomponents. Thus, a hierarchically modeled component is really a subsystem.

We use the term *component* to refer to a portion of a system that can be tested and replaced as a unit, whether or not it can be decomposed into subcomponents. To specify a component hierarchically, the user has to specify the component's input variables, the output variable, and the mode variable for the component (as in the case of an atomic model). Specifying a variable involves identifying its name and its states. In addition, the user has to specify a *subcomponent model* for the component. A subcomponent model is simply another hierarchical system model. The system input variables of the subcomponent model are the same as the input variables of the component and the system output variable of the subcomponent model is the the output variable of the component. Note that each subcomponent in the subcomponent model can itself be modeled hierarchically or atomically.

In repairing hierarchically structured systems, we must consider two costs for each component $C$—the replacement cost and the inspection cost. The replacement cost $c$ is the cost of replacing the entire component. The inspection cost $d$ "buys access" to the values of the component's input variables and output variable during the repair process. If we decide to pay the inspection cost and then find that the observed output reading of the component is anomalous for the observed input, we have two options: (1) replace the entire component, or (2) successively repair subcomponents of the component, checking whether the anomalous output has been fixed after each subcomponent repair.

As in Section 2.1, we assume that the input value remains fixed. We also assume that a hierarchically modeled component is in the **ok** state if and only if every one of its subcomponents is in the **ok** state. Any joint state of the subcomponents which includes any non-**ok** state thus has to correspond to some non-**ok** state of the component. This correspondence is specified as an *abstraction function* (see [Srinivas, 1994]).

For simplicity, in this presentation we will assume that each component has only two possible states—normal (**ok**) and broken (**b**). In this case, every possible joint state of the subcomponents which contains at least one **b** state will automatically map to the **b** state of the containing component.

### 3.2 COMPILING THE HIERARCHICAL MODEL

Say a component $C$ with inputs **I**, output $O$ and mode variable $M$ is modeled hierarchically, and its subcomponents $C_i^s$ are modeled atomically. We can easily compute the distribution $P(M)$. We do this as $P(M = \mathbf{ok}) = \Pi_i P(M_i^s = \mathbf{ok}_i)$ and $P(M = \mathbf{b}) = 1 - P(M = \mathbf{ok})$. In addition, we can sum out all the internal variables of the subcomponent model to obtain the distribution $P(O|\mathbf{I}, M)$ (see [Srinivas, 1994]). Thus, we can compile the subcomponent model into an *atomic description* of the component. As we shall see, we will use these atomic descriptions when computing the optimal repair plan.

It is not necessary that the subcomponents $C_i^s$ be modeled atomically for this compilation process to be possible. We need only to ensure that an atomic description of each $C_i^s$ is available when computing the atomic description of $C$. This can be ensured by a bottom up traversal of the hierarchy tree of the system during which each component's atomic description is computed.

### 3.3 HIERARCHICAL REPAIR PLANS

A component's *hierarchical repair plan* specifies a *repair strategy* for each joint value **i** of the input variables **I** of the component. The repair strategy specifies what needs to be done if the input value is **i** and the output is anomalous. The repair strategy can either specify (a) replacement of the component or (b) repair of subcomponents. We will make the assumption that if an anomaly in a component is fixed by repair of subcomponents, this results in the component being returned to the *ok* state (as in the case of replacement).

When a strategy calls for repair of subcomponents, it also specifies an order in which the subcomponents are to be successively repaired until the output is no longer anomalous. In addition, the strategy also specifies a *repair method* for each subcomponent. The *repair method* for the subcomponent can be either (a) replacement (without inspection) or (b) inspection followed by repair (this repair is according to the subcomponent's hierarchical repair plan). The specification of a hierarchical plan for a component also includes a specification of a hierarchical repair plan for each of its subcomponents.

An optimal hierarchical repair plan for a component specifies the strategy of least expected cost for each joint input value **i**. At the top level of the hierarchical system, the entire system can be viewed as a single component. The optimal hierarchical repair plan for a system is simply the optimal hierarchical repair plan for the component.

## 3.4 COMPUTING THE OPTIMAL HIERARCHICAL REPAIR PLAN

Consider a component $\mathcal{C}$ with inputs $\mathbf{I}$, output $O$ and mode variable $M$ which is modeled hierarchically. Let the component's replacement cost be $c$. Assume it has subcomponents $\mathcal{C}_1^s, \mathcal{C}_2^s, \ldots, \mathcal{C}_n^s$, each of which has an atomic description available.

Consider computing the optimal strategy for repair of $\mathcal{C}$ given that the input has the value $\mathbf{i}$ and the output is anomalous. Let us suppose that the subcomponents cannot be inspected. In this case, the only repair available for each subcomponent $\mathcal{C}_i^s$ is to replace it, incurring cost $c_i^s$. We note that we can use the algorithm developed in Section 2.1 to compute the optimal sequence $OptSeq(\mathbf{i})$ in which to replace the components for this particular input. Let the expected cost of this sequence be $EC(OptSeq(\mathbf{i}))$.

We note that if $EC(OptSeq(\mathbf{i})) \geq c$ then the optimal strategy for input $\mathbf{I} = \mathbf{i}$ is simply to replace component $\mathcal{C}$ if the output is anomalous. In the case where $EC(OptSeq(\mathbf{i})) < c$ then the optimal strategy for input $\mathbf{I} = \mathbf{i}$ is to replace the components successively in the order $OptSeq(\mathbf{i})$.

Now, let us consider the general situation where we have the choice of either replacing a subcomponent or inspecting it and then repairing it according to its hierarchical repair plan. Assume that each subcomponent's hierarchical repair plan has already been computed and stored. Given a particular input $\mathbf{i}$ and some sequence $Seq$ in which the subcomponents are to be repaired, we will develop a algorithm (Section 3.4.1) that chooses the repair method for each subcomponent such that the expected cost of repair is minimized. The algorithm outputs the sequence $Seq$ annotated with the best repair method for each component. Let us call this annotated sequence $Seq^m$. The algorithm also outputs the expected cost $EC(Seq^m)$ of this optimal strategy.

If we now compare $EC(Seq^m)$ across every possible sequence $Seq$ we can find the best repair sequence (and the accompanying repair methods) for the input $\mathbf{i}$. Let this best sequence be $OptSeq^m(\mathbf{i})$. We can choose the best strategy of repair for $\mathcal{C}$ given the input is $\mathbf{i}$ and the output is anomalous as follows. If $c \leq EC(OptSeq^m(\mathbf{i}))$ then the optimal strategy is simply to replace component $\mathcal{C}$. If $c > EC(OptSeq^m(\mathbf{i}))$, then the optimal strategy is to repair subcomponents in the sequence (and with the methods) specified by $EC(OptSeq^m(\mathbf{i}))$.

An optimal hierarchical repair plan for the hierarchical system can be computed through a bottom-up traversal of the tree representing the hierarchical decomposition of the system. At each node of the tree (representing each component), the algorithm described above can be used to synthesize an optimal hierarchical repair plan for the component from the subcomponent model, as well as the optimal hierarchical repair plans of the subcomponents.

### 3.4.1 Computing $Seq^m$

Let us consider the case where a system component $\mathcal{C}$ has $n$ subcomponents. Let $X$ be the component's output variable. Assume that the optimal hierarchical repair plan for each subcomponent $\mathcal{C}_j^s$ of $\mathcal{C}$ has already been computed and is available. Let $OptEC_j^s(\mathbf{I}_j^s = \mathbf{i}_j^s, X_j^s = \neg x_j^s(\mathbf{i}_j^s))$ be the optimal repair cost of $\mathcal{C}_j^s$ given that its input is $\mathbf{i}_j^s$ and that its output $X_j^s$ is anomalous. Note that this repair cost is available from the precomputed optimal hierarchical plan for the subcomponent $\mathcal{C}_j^s$.

Let $Seq^a$ be some annotated sequence in which the components are to be repaired. The annotation specifies a repair method $m_j$ for each subcomponent $\mathcal{C}_j^s$. If $m_j = $ **replace** then $\mathcal{C}_j^s$ is simply replaced. If $m_j = $ **inspect** then $\mathcal{C}_j^s$ is inspected and then repaired according to its optimal hierarchical repair plan. The cost of repair of $\mathcal{C}_j^s$ in $Seq^a$ is a function of the method of repair $m_j$. We refer to this cost as $Cost(\mathcal{C}_j^s, m_j)$.

We can compute the expected cost of $Seq^a$ as:

$$EC(Seq^a | \mathbf{I} = \mathbf{i}, X = \neg x(\mathbf{i})) = \qquad (5)$$
$$(Cost(\mathcal{C}_1^s, m_1) +$$
$$P(X_1 = \neg x_\mathbf{i} | S_1)(Cost(\mathcal{C}_2^s, m_2) +$$
$$P(X_2 = \neg x_\mathbf{i} | S_2)(Cost(\mathcal{C}_3^s, m_3) +$$
$$\ldots$$
$$P(X_{j-1} = \neg x_\mathbf{i} | S_{j-1})(Cost(\mathcal{C}_j^s, m_j)$$
$$\ldots$$
$$P(X_{n-1} = \neg x_\mathbf{i} | S_{n-1}) Cost(\mathcal{C}_n^s, m_n) \ldots) \ldots)))$$

The probabilities $P(X_j = \neg x_\mathbf{i} | S_j)$ for $1 \leq j \leq n$ can computed as described earlier with the iterative algorithm of Section 2.1. However, we have yet to specify how to compute $Cost(\mathcal{C}_j^s, m_j)$.

If $m_j = $ **replace** then $Cost(\mathcal{C}_j^s, m_j)$ is simply the replacement cost $c_j^s$ of $\mathcal{C}_j^s$. If $m_j = $ **inspect** then the cost has two components. The first is the inspection cost. The second is the expected cost of fixing $\mathcal{C}_j^s$ hierarchically. This expected cost *depends on the current context*. The current context includes the observations $S_{j-1}$ and the observation $X_j = \neg x(\mathbf{i})$. Thus we have $Cost(\mathcal{C}_j^s, m_j) = d_j^s + ECHR(\mathcal{C}_j^s | S_{j-1}, X_j = \neg x(\mathbf{i}))$. Here, $ECHR$ is the expected cost of repair of $\mathcal{C}_j^s$ using its optimal hierarchical repair plan in the current context.

The current context gives us updated information about the probability distribution over the input and output of $\mathcal{C}_j^s$. This updated information is summarized by the conditional distribution $P(\mathbf{I}_j^s, X_j^s | S_j, X_j = \neg x(\mathbf{i}))$. Given this conditional distribution, $ECHR(\mathcal{C}_j^s | S_{j-1}, X_j = \neg x(\mathbf{i}))$ can be computed as:

$$ECHR(\mathcal{C}_j^s | S_{j-1}, X_j = \neg x(\mathbf{i})) =$$
$$\Sigma_{\mathbf{i}_j^s} OptEC_j^s(\mathbf{I}_j^s = \mathbf{i}_j^s, X_j^s = \neg x_j^s(\mathbf{i}_j^s)) \times$$
$$P(\mathbf{I}_j^s = \mathbf{i}_j^s, X_j^s = \neg x_j^s(\mathbf{i}_j^s) | S_{j-1}, X_j = \neg x(\mathbf{i}))$$

Note that the second term within the summation is just the posterior probability of seeing input $\mathbf{i}_j^s$ in conjunction with an anomalous output. When this situation occurs we use the optimal hierarchical strategy of cost $OptEC_j^s(\mathbf{I}_j^s = \mathbf{i}_j^s, X_j^s = \neg x_j^s(\mathbf{i}_j^s))$. Thus, the above equation computes the expected cost. The problem now reduces to that of computing the distribution $P(\mathbf{I}_j^s, X_j^s | S_j, X_j = \neg x(\mathbf{i}))$. We will solve this below.

In summary, we now have an algorithm for calculating the expected cost of the annotated sequence $Seq^a$. This algorithm can be modified to give the optimal annotated sequence $Seq^m$ as follows: Given a sequence $Seq$, if $Cost(\mathcal{C}_j^s, \textbf{replace}) < Cost(\mathcal{C}_j^s, \textbf{inspect})$, we choose to replace rather than inspect $\mathcal{C}_j^s$. If the inequality is reversed, we choose to inspect rather than to replace the component.

The distribution $P(\mathbf{I}_j^s, X_j^s | S_j, X_j = \neg x(\mathbf{i}))$ can be computed using the same idea which was used to compute $P(X_j = \neg x(\mathbf{i}) | S_j)$ in Section 2.1. That is, we can use the fact that knowing the state (mode) of every component and the system input makes $\mathbf{I}_j^s$ independent of the history of observations and repairs $S_j$. Let $M_{[i,j]}$ represent the set of variables $\{M_i, M_{i+1}, \ldots, M_j\}$. Let $m_{[i,j]}$ be a generic joint state of these variables. Let $\mathbf{ok}_{[i,j]}$ be the state where each of the variables in the $\mathbf{ok}$ state.

We can then compute the needed probability distribution as:

$$P(\mathbf{I}_j^s = \mathbf{i}_j^s, X_j^s = x_j^s | S_j, X_j = \neg x(\mathbf{i})) = \quad (6)$$
$$\Sigma_{\mathbf{m}_{[j+1,n]}} P(\mathbf{I}_j^s = \mathbf{i}_j^s, X_j^s = x_j^s | \mathbf{I} = \mathbf{i}, \mathbf{M}_{[1,j]} = \mathbf{ok}_{[1,j]},$$
$$\mathbf{M}_{[j+1,n]} = \mathbf{m}_{[j+1,n]}, X_j = \neg x(\mathbf{i})) \times$$
$$P(\mathbf{M}_{[j+1,n]} = \mathbf{m}_{[j+1,n]} | S_j)$$

The first term in the summation can be computed directly from the static Bayesian network corresponding to the model. Consider applying the clustering inference algorithm to the static network ([Lauritzen & Spiegelhalter, 1988; Jensen et al, 1990]). The clustering inference algorithm guarantees that for each component $\mathcal{C}_j^s$, the output variable $X_j^s$ and the variables in $\mathbf{I}_j^s$ necessarily will be in the same clique. This is because each variable in $\mathbf{I}_j^s$ is a parent of $X_j^s$. Let this clique be $Clique_j^s$. If we propagate the evidence $\mathbf{i}$, $\mathbf{ok}_{[1,j]}$, $\mathbf{m}_{[j+1,n]}$ and $X = \neg x(\mathbf{i})$ in this network, and then sum over all the variables other than $\mathbf{I}_j^s$ and $X_j^s$ in the posterior belief of $Clique_j^s$ and renormalize, we obtain the distribution $P(\mathbf{I}_j^s, X_j^s | \mathbf{I} = \mathbf{i}, \mathbf{M}_{[1,n]} = \mathbf{m}_{[1,n]}, X_j = \neg x(\mathbf{i}))$. Note that the computation of $P(\mathbf{I}_j^s = \mathbf{i}_j^s, X_j^s = x_j^s | S_j, X_j = \neg x(\mathbf{i}))$ can be integrated into the iterative algorithm of Section 2.1.

The distribution $P(\mathbf{M}_{[j+1,n]} = \mathbf{m}_{[j+1,n]} | S_j)$ in the second term of the above equation is computed by the iterative algorithm of Section 2.1.

### 3.4.2  Repair Algorithm: Review

In review, the algorithm for computing the optimal hierarchical repair plan proceeds as follows:

1. **Model compilation**: An atomic description for each component is computed by a bottom up traversal of the hierarchy tree.

2. **Plan computation**: In a bottom up traversal of the hierarchy tree, for each component $\mathcal{C}$ with output $O$ and input $\mathbf{I}$ and subcomponents $\mathcal{C}_j^s$, $1 \leq j \leq n$:

    For every possible input $\mathbf{i}$ of $\mathcal{C}$:
    - For every possible sequence $Seq$ of the subcomponents, compute the optimal annotation $Seq^m$ and its expected cost. Let the cheapest of these annotated sequences be $OptSeq^m(\mathbf{i})$.
    - Compare the cost of $OptSeq^m(\mathbf{i})$ with the replacement cost $c$ to compute the optimal repair strategy for input $\mathbf{i}$.

### 3.5  COMPLEXITY OF THE ALGORITHM

Say that any variable in the hierarchical system model (input, output or mode) has at most $s$ states, and that $s$ is small. This is reasonable for discrete systems with a small number of states for any state variable (e.g., such as digital circuits). Let the maximum number of subcomponents of any component be $b$—this is also the branching factor of the tree representing the system hierarchy. If each component has at most $m$ input variables, the number of variables in the subcomponent model is $O(m \times b)$.

It is straightforward to show that the computation of the optimal hierarchical repair algorithm at every node of the hierarchy tree is bounded by $O(b! s^{(m \times b)})$. If we assume that the $b$ and $m$ are small (which is reasonable if we are modeling a system hierarchically) the computation is tractable (see Section 3.7). Let us refer to the expression $(b! s^{(m \times b)})$ as $B$.

If a hierarchical model has $N$ leaf level components in the hierarchy tree, then the total number of components (including interior nodes in the tree) is $O(N)$. The optimal hierarchical repair algorithm for this model has an overall complexity of $O(NB)$. Hence, the performance of the algorithm is linear in the size of the hierarchical system. The algorithm is thus tractable if $B$ is reasonably small.

## 3.6  INTRODUCING FLEXIBILITY

We can modify the algorithm to flexibly trade off the quality of repair plans with computation time. Flexible, incremental-refinement procedures make it possible to maximize the value of computational procedures given variation and uncertainty in the costs of time [Horvitz, 1988]. Consider a hierarchy tree with $d$ levels of branching and a branching factor of $b$. The top (first) level of the hierarchy tree has one component, the next (second) level has $b$ components, the third level has $b^2$ components and so on till the $d$-th level which has $b^{d-1}$ components. Say that we apply the exhaustive algorithm for computing the optimal hierarchical repair plan repeatedly, but with the following modification: At the $k$-th iteration, we assume that the components at the $k$-th level *cannot* be inspected – they can only be replaced. As a result, in the $k$-th iteration we need to consider only those components which are between level 1 and level $k$ when computing the optimal hierarchical plan.

Thus, in the first iteration, the modified algorithm considers only replacement of the entire system if the output is anomalous. In the second phase, the method compares replacement of the entire system with inspecting it and carrying out an optimal replacement sequence of its direct subcomponents – the components at the second level of the hierarchy tree. The third iteration considers inspection of the subcomponents at the second level of the hierarchy tree but only replacement of components at the third level of the hierarchy tree, and so on. The $k$-th iteration of the algorithm will compute an optimal hierarchical plan in $O(b^k B)$. Thus early iterations run very quickly but give crude answers. Later iterations take longer but give more refined answers. Note that the iterative process can be interrupted at any time to yield the current answer. Essentially, we are applying the exhaustive algorithm with iterative deepening.

## 3.7  IMPLEMENTATION

We have implemented the algorithm in LISP on a Sun workstation. Running on a hierarchical digital circuit with a branching factor $b = 4$ and $m = 3$, the (unoptimized) implementation takes about 3.2 minutes to compute the policy at each node of the hierarchy tree. For a system with 256 leaf level components this amounts to about 18 hours of computation to compute the global optimal hierarchical plan. Thus, the present implementation can be considered suitable for off-line precomputation of optimal hierarchical repair plans for medium sized systems. Of course, if we employ the iterative-deepening version of the algorithm, we can get lower quality answers more quickly. For example, if the flexible repair strategy looks only 3 levels down the hierarchy tree, the optimal plan can be computed in 67 minutes. We expect that an optimized implementation of the algorithm should be able to scale up to systems with thousands of components.

## 4  DISCUSSION

Computing an optimal repair strategy is intractable for a general formulation of the repair problem (see for example, [Heckerman *et al*, 1995]). In the general case, we must identify the best repair strategy from a combinatorial space of strategies. Two types of simplifications have been employed to address this complexity. With one approach, the repair problem is restricted by assuming or identifying additional structure in the problem. In the other approach, attempts are made to compute some good immediate repair action, based on a limited lookahead. Such greedy approaches are applied in an interactive procedure, where a recommended action is used to gather additional information that is used in the next iteration of the myopic analysis. We have taken the first of these approaches— we exploit the system hierarchy to get computational gains.

Let us explore in more detail the basis for the efficiencies we gain by exploiting hierarchical structure. Consider a hierarchical system where $C_1^l$ and $C_2^l$ are subcomponents of $C_1$ and $C_3^l$ and $C_4^l$ are subcomponents of $C_2$. Say the inspection costs of non-leaf components is zero, that their replacement cost is very high, and that leaf components cannot be inspected. In such a case, the optimal repair sequence is a sequence of leaf component replacements. If we ignored hierarchy, we would have to consider every possible sequence of the leaf components. However, when we represent and exploit the hierarchy, some of these sequences are impossible (for example, $\langle C_1^l, C_3^l, C_2^l, C_4^l \rangle$). This is because the repair protocol specifies that either $C_1$ is fixed first or $C_2$ is fixed first. Thus we are effectively considering only those repair sequences in which both of $C_1^l$ and $C_2^l$ appear (in some order) before $C_3^l$ and $C_4^l$ (in some order) and those sequences in which both of $C_3^l$ and $C_4^l$ appear before $C_1^l$ and $C_2^l$. Thus, the hierarchy gives us a substantial reduction in the search space.

The approach we have taken *takes advantage* of the system model, rather than restricting it. Other researchers have shown how restricting the system model can make precomputation of repair strategies tractable. [Kalagnanam & Henrion, 1988] derive an optimality condition for the optimal repair strategy in a multi-component system which is assumed to have a single fault. The repair protocol is similar to the one described in this paper with the exception that only component replacements are allowed. There is no notion of inspection of components. [Srinivas, 1995] generalizes the result of [Kalagnanam & Henrion, 1988] to the case of multiple independent failures and introduces a formulation of component inspection. [Heckerman *et al*, 1995] also employs the single-fault restriction. In this work, repair is formulated as an interactive process. The system is modeled with a Bayesian network and both component replacement and information gathering actions are possible. An action is chosen at each step of the process, taking advantage of a myopic heuristic.

The work in the model-based diagnosis community ([Hamscher et al, 1992]) has also addressed the repair problem as an interactive process. [deKleer & Williams, 1987] introduce an entropy-based method for observation planning. [Friedrich & Nejdl, 1992] develop a set of greedy algorithms for choosing observation and repair actions in interactive model-based diagnosis. Their approach explicitly considers downtime costs of unanticipated failures. Hence their repair scheme implicitly includes a notion of preventive maintenance. [Poole & Provan, 1991] use repair actions to partition the world into a set of classes. All the worlds in a class result in the same action response. In their formulation, the diagnosis problem becomes one of determining the class of the current state of the system. [Yuan, 1993] proposes a decision-theoretic framework for modeling interactive model-based diagnosis. At each step of the diagnosis, a decision model, in the form of an influence diagram, is synthesized and solved to compute the next action. The model is successively refined along the system hierarchy using a single fault assumption until the fault is located.

The emphasis in this paper has been to address the problem of precomputing good strategies rather than interleaving action and repair planning. We have presented the hierarchical repair algorithm developed in this paper in the context of model-based diagnosis. However, it is equally applicable to diagnosis models which are developed directly as hierarchical Bayesian networks.